\documentclass[letterpaper, 10pt, conference]{ieeeconf}  
\IEEEoverridecommandlockouts                              % This command is only needed if 
\usepackage{units}
\usepackage{bm}
\usepackage{float}
\usepackage[font=small,skip=7pt]{caption}
\usepackage[ruled,vlined,noend]{algorithm2e}
\usepackage{hyperref}
\usepackage{microtype}

\usepackage{epsfig}
\newlength\fwidth
\usepackage[T1]{fontenc}
\usepackage{mathtools}

\usepackage{xcolor}
\usepackage{siunitx}
\usepackage{multirow}
\usepackage{epstopdf}
\usepackage{tabularx}
\usepackage{amsmath} % assumes amsmath package installed
\usepackage{amssymb}  % assumes amsmath package installed
\usepackage{mathtools}
\DeclareGraphicsExtensions{.pdf,.eps}
\DeclareMathOperator*{\argmin}{arg\,min}
\usepackage[acronym,nomain,nonumberlist]{glossaries}
\newcommand{\N}{\mathbb{N}}

\newcommand\smashunderbracket[3][0.5pt]{%
	\vphantom{#2}%
	\smash{\underbracket[#1]{#2}_{#3}}%
}

\newacronym{mav}{MAV}{Micro Aerial Vehicle}
\newacronym{nmpc}{NMPC}{Nonlinear Model Predictive Control}
\newacronym{mpc}{MPC}{Model Predictive Control}
\newacronym{panoc}{PANOC}{Proximal Averaged Newton-type method for Optimal Control}

\usepackage{url}
\title{\LARGE \bf Exploration-RRT: A multi-objective Path Planning and Exploration Framework for Unknown and Unstructured Environments. %\thanks{This work has been partially funded by the European Unions Horizon 2020 Research and Innovation Programme under the Grant Agreement No. 730302 SIMS.}
}

 \author{Bj\"orn Lindqvist$^1$, Ali-akbar Agha-mohammadi$^2$ and George Nikolakopoulos$^1$.
  
\thanks{$^{1}$The authors are with the Robotics and Artificial Intelligence Team, Department of Computer, Electrical and Space Engineering, Lule\r{a} University of Technology, Lule\r{a} SE-97187, Sweden. Corresponding Author's email: \texttt{bjolin@ltu.se.}}%
\thanks{$^{2}$The author is with the Jet Propulsion Laboratory California Institute of Technology Pasadena, CA, 91109.}
\thanks{This work has been partially funded by the European Unions Horizon 2020 Research and Innovation Programme under the Grant Agreement No. 869379 illuMINEation.} 
}
\begin{document}
\captionsetup{font=footnotesize}
\maketitle
\thispagestyle{empty}
\pagestyle{empty}
%%%%%%%%%%%%%%%%%%%%%%%%%%%%%%%%%%%%%%%%%%%%%%%%%%%%%%%%%%%%%%%%%%%%%%%%%%%%%%%%
\begin{abstract}
This article establishes the Exploration-RRT algorithm: A novel  general-purpose combined exploration and path planning algorithm, based on a multi-goal Rapidly-Exploring Random Trees (RRT) framework. Exploration-RRT (ERRT) has been specifically designed for utilization in 3D exploration missions, with partially or completely unknown and unstructured environments. The novel proposed ERRT is based on a multi-objective optimization framework and it is able to take under consideration the potential information gain, the distance travelled, and the actuation costs, along trajectories to pseudo-random goals, generated from considering the on-board sensor model and the non-linear model of the utilized platform. In this article, the algorithmic pipeline of the ERRT will be established and the overall applicability and efficiency of the proposed scheme will be presented on an application with an Unmanned Aerial Vehicle (UAV) model, equipped with a 3D lidar, in a simulated operating environment, with the goal of exploring a completely unknown area as efficiently and quickly as possible. 
\end{abstract}
\glsresetall %it resets the abbreviations to be redefine in the introduction
%%%%%%%%%%%%%%%%%%%%%%%%%%%%%%%%%%%%%%%%%%%%%%%%%%%%%%%%%%%%%

%%%%%%%%%%%%%%%%%%%%%%%%%%%%%%%%%%%%%%%%%%%%%%
\section{Introduction and Background}
%%%%%%%%%%%%%%%%%%%%%%%%%%%%%%%%%%%%%%%%%%%%%%
The exploration and mapping of unknown and complex structures has been always one of the main application areas of autonomous robots \cite{almadhoun2016survey}, including the inspection of large-scale infrastructures\cite{mansouri2018cooperative}, search-and-rescue missions\cite{beck2016online}, and subterranean exploration \cite{kanellakis2018towards}. Specifically, the area of subterranean environments, and in general harsh and GPS-denied ones, has been in focus for autonomous exploration missions with the utilization of autonomous robots. This area has gained a lot of attention in the latest years, especially due to the DARPA subterranean challenge\cite{subt}, where teams of robots are tasked to explore and identify artifacts in complex and unstructured environments. Together with Simultaneous Localization and Mapping (SLAM), the problem of path planning and exploration behavior\cite{palieri2020locus, kim2021plgrim} is one of the most fundamental problems in such applications, which was the main inspiration for this work as part of the NEBULA autonomy developments\cite{nebula}. 

\begin{figure}[!htb]
    \centering
\includegraphics[width=\columnwidth]{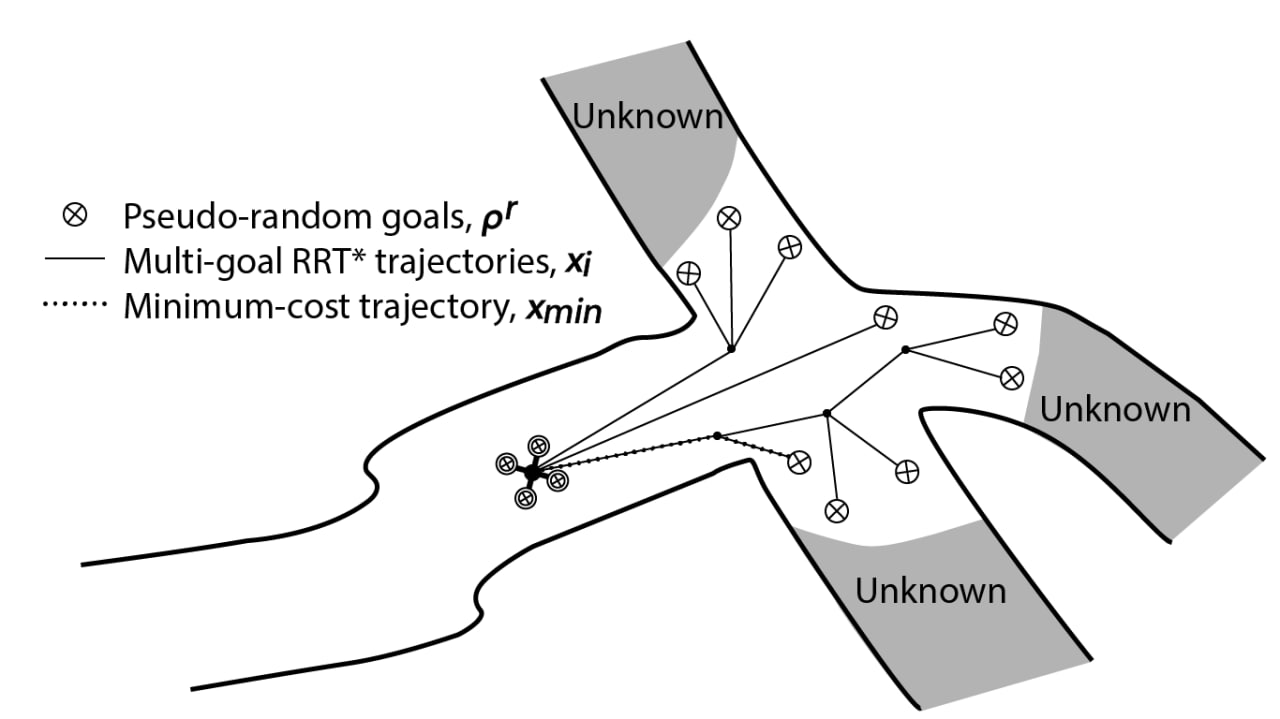}
    \caption{The ERRT concept. Multiple trajectories are calculated to pseudo-random goals, and the lowest cost trajectory is chosen.}
    \label{fig:concept}
\end{figure}

Both path planning and exploration have been cornerstone research areas in the field of robotics and have been massively studied as for example in~\cite{zhao2018survey},~\cite{yang2016survey},~\cite{amigoni2017multirobot},~\cite{julia2012comparison}. %SURVEYS
At the same time, it should be mentioned that there is a very large range of solutions to the problem of total exploration of unknown areas, including pattern-based solutions\cite{gonzalez2005bsa}, frontier\cite{zhu20153d} or entropy\cite{carrillo2015autonomous} approaches that have been also extensively demonstrated. Occupancy-based path planning (e.g. path planning that relies on a map of occupied and free space) are fundamentally based either on applications of Djikstra's Algorithm\cite{noto2000method}, with modern examples including improved versions of $\text{A}^*$\cite{chaari2017design} or Jump-point-search\cite{duchoe2014path}, or on the Rapidly-Exploring Random Tree (RRT) algorithm \cite{lavalle1998rapidly}, that also has numerous moderns improved versions \cite{adiyatov2017novel}\cite{otte2016rrtx} and is the core component of the proposed path planner as well. The advantage of RRT is its computational efficiency and the ease of adding functionalities\cite{ma2015efficient} on top of the central planning problem, while its downside is that for limited iterations there is no guarantee of finding the shortest path to the goal. 

In many applications, the two problems of exploration and planning are solved separately. In \cite{niroui2019deep} a Deep Reinforcement model selects optimal frontiers and an A$^*$ algorithm is applied to find the shortest path. In \cite{shen2012autonomous} stochastic differential equations are used for identifying optimal frontiers, while a RRT$^*$ path planner computes the path.

Incorporating an exploration behavior into the central planning problem is a research area that gained significant attention lately, with multiple proposed solutions\cite{dharmadhikari2020motion}\cite{wang2020actively}, where the Next-Best-View planners\cite{pito1999solution} established the fundamental concept. The flagship modern Next-Best-View planners have seen multiple application scenarios as local path planners \cite{bircher2016receding}\cite{dang2020autonomous} to name a few. The core concepts of Exploration-RRT (ERRT) are in the same direction as Next-Best-View planners, while its novelty comes from the difference in the algorithmic implementation where ERRT is explicitly solving and evaluating paths to pseudo-random goals versus an iterative evaluation of RRT-branches until a sufficiently good branch is found. ERRT novelty is also supported from the ability to compute the optimal actuation required to track the generated trajectory, which is done via solving a receding horizon NMPC problem. In the proposed method the NMPC problem is solved by the Optimization Engine\cite{sopasakis2020open} an open-source and very fast Rust-based optimization software. 
%
%%%%%%%%%%%%%%%%%%%%%%%%%%%%%%%%%%%%%%%%%%%%%%
\section{Contributions}
%%%%%%%%%%%%%%%%%%%%%%%%%%%%%%%%%%%%%%%%%%%%%%
%
This article proposes a novel solution to the combined path planning and exploration problem, based on a minimum-cost formulation problem with pseudo-random generated goals combined with multi-path planning and evaluation. The proposed method evaluates the model-based actuation based on a nonlinear system model along each generated trajectory, which, to the authors best knowledge, has never been included in such a scheme before, together with the information gain and the total distance. Additionally, ERRT considers the full coverage of the unknown map, as long as feasible frontiers or points-of-interest exist. We evaluate the scheme specifically for an UAV platform with an added 3D lidar sensor model, analysing the consistency and efficiency of exploring a completely unknown subterranean-like area. The algorithm is designed around a very general input-output model that allows for the user to on-the-fly change the desired goal and behavior of the planner.
%%%%%%%%%%%%%%%%%%%%%%%%%%%%%%%%%%%%%%%%%%%%%%
\section{Methodology} \label{sec:methodology}
%%%%%%%%%%%%%%%%%%%%%%%%%%%%%%%%%%%%%%%%%%%%%%

%%%%%%%%%%%%%%%%%%%%%%%%%%%%%%%%%%%%%%%%%%%%%%
\subsection{The Problem}
The overarching goal of coupling the path planning and exploration problem is considered in the proposed ERRT framework as the minimization of three quantities namely: The total distance of the 3D trajectory $\bm{x}$, the actuation required to track the trajectory, where $\bm{u}$ denotes a series of control actions, and increasing the known space, here considered as a negative cost associated with the information gain $\nu\in\mathbb{R}$ along trajectory $\bm{x}$. These quantities result in the following minimization:
%\begin{subequations}\label{eq:realproblem}
\begin{align}\label{eq:realproblem}
\operatorname*{Minimize} J_a(\bm{u}) + J_d(\bm{x}) + J_e(\nu) \\
\text{subj. to:}\,& \bm{x} \in V_{free} \notag \\
& J_e(\nu) \neq 0 \notag
\end{align}
%\end{subequations}
with $J_a(\bm{u})$ denoting the actuation cost, $J_d(\bm{x})$ the distance cost, $J_e(\nu)$ the exploration, or information-gain cost, which should be non-zero to expand the known space, and $V_{free} \in V_{map}$ denoting the obstacle-free space, with $V_{map} \in \mathbb{R}^3$ as the 3D position-space encompassed by the current map configuration. If solved completely, this would result in the optimal trajectory for the exploration task, in a compromise between quickly discovering more space, limiting actuation based on a dynamic system model, and being \textit{lazy} e.g. moving as little as possible. 

\subsection{Proposed Solution}
\begin{figure*}[ht]
    \centering
\includegraphics[scale=0.32]{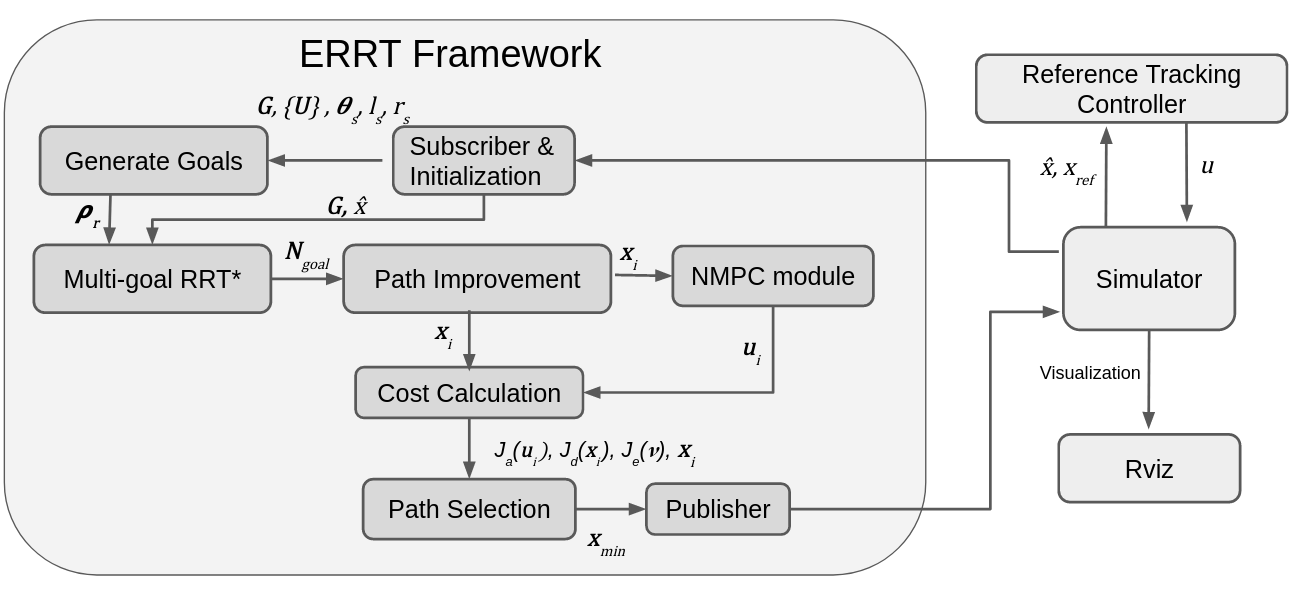}
    \caption{Exploration-RRT framework and pipeline (left) and simulation architecture (right).}
    \label{fig:structure}
\end{figure*}

Towards this goal, ERRT proposes a solution composed of four components: pseudo-random goal generation based on a sensor model, a multi-goal RRT$^*$ planner, a receding horizon optimization problem (Nonlinear MPC) to solve for the optimal actuation along the trajectory, and finally computing the total costs associated with each trajectory and choosing the minimal-cost solution. An overview of this process is found in Figure \ref{fig:structure}. Each component will be explained in more detail in \ref{sec:algorithm}. 
The process of generating many goals, solving the path to each of them, and then computing the total costs associated with each trajectory, turns~\eqref{eq:realproblem} from a true optimization problem into: 
%\begin{subequations}
\begin{align}\label{eq:approx}
    \argmin (J_a(\bm{u}_i) + J_d(\bm{x}_i) + J_e(\nu_i))_i,  i = 1, \ldots, n_{goals} \\
    \text{where}~ \bm{x}_i \in V_{free}\notag
\end{align}
%\end{subequations}
that considers the finding of $\bm{x}_{\mathrm{min}} \in V_{free}$
%\in \mathbb{R}^3$% 
that is the $\bm{x}_i$ trajectory having the lowest cost associated to it
%, where the first and last element of $\bm{x}_i$ are the initial and the \textit{i:th} goal positions respectively
, and $n_{goal}\in \mathbb{N}$ is the overall number of path planner goals. This process will converge towards approximating the complete problem in~\eqref{eq:realproblem}, as we increase $n_{goal}$ and optimize the trajectory-generating algorithm (or simply by increasing the number of iterations of the RRT$^*$), as more and more possible solutions are being investigated. 
\subsection{The Algorithm}\label{sec:algorithm}
For the sake of notation, let us directly define desired points to explore, or unknown voxels, as $\{U\}$, and a binary 3D grid map $\bm{G}$ where occupied voxels, $\{O\}$ and unknown voxels $\{U\}$ are set to 1, and the resulting $V_{free}$ is set to 0. Let us also define the measured vehicle state as $\hat{x}$.

\subsubsection{Goal Generation}\label{sec:goals}
\begin{figure}[!htb]
    \centering
\includegraphics[width=0.9\columnwidth]{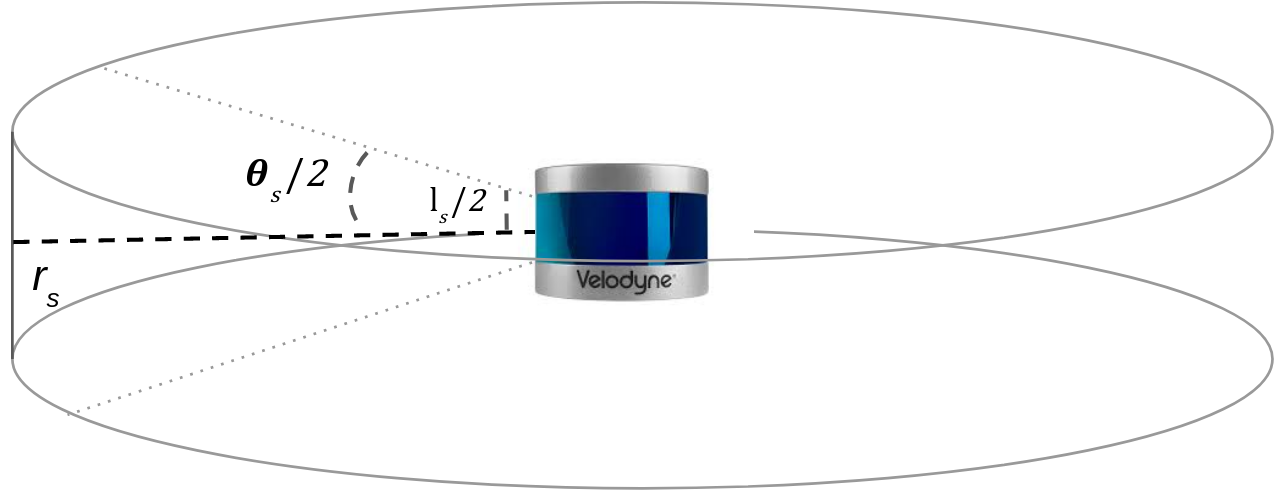}
    \caption{Simplified sensor model of a 3D lidar. $r_s$ denotes the sensor range and $\theta_s$ the field-of-view. $l_s$ denotes the size of the inner scanner array.}
    \label{fig:sensormodel}
\end{figure}

The fundamental part of ERRT that allows for the exploration behavior, is the generation of pseudo-random goals $\rho^r$ within $V_{map}$. At each call to the algorithm, $n_{goal}$ goals are generated, under the conditions of being inside the sensor range of at least one unknown voxel $U\in \{U\}$. Other conditions are $\rho^r \in V_{free}$, and the straight path from $\rho^r$ to the center of $U$ being obstacle-free (including being blocked by other unknown voxels). While many works focus on onboard cameras~\cite{bircher2016receding} ERRT considers an onboard 3D lidar, with the advantage in terms of mapping being a \unit[360]{$^{\circ}$} vision in the \textit{x-y} plane and a long sensor range, and with the drawback of a narrow field-of-view in \textit{z} close to the platform, thus making narrow 3D spaces a challenge. A simplified sensor model of a 3D lidar is shown in Figure \ref{fig:sensormodel} assuming the lidar is placed flat on the platform, considering only three parameters: the sensor range $r_s\in\mathbb{R}$, the field-of-view $\theta_s\in\mathbb{R}$ and $l_s\in\mathbb{R}$, describing the size of the sensor array. 
\\Assuming a randomly generated point $\rho^r = [\rho^r_x, \rho^r_y, \rho^r_z]\in\mathbb{R}^3$ and a center-position of an unknown voxel $U$ as $p^u = [p^u_x, p^u_y, p^u_z]\in\mathbb{R}^3$, we can evaluate, by assuming small roll/pitch angles in the UAV case, if $\rho^r$ is inside the space seen by the sensor model if $r_s \geq \sqrt{(\rho^r_x - p^u_x)^2 + (\rho^r_y - p^u_y)^2}$ and $\mid \rho^r_z - p^u_z \mid  \leq \sqrt{(\rho^r_x - p^u_x)^2 + (\rho^r_y - p^u_y)^2}\tan(\frac{\theta_s}{2}) + \frac{l_s}{2}$. This results in Algorithm 1
\begin{algorithm}[ht] 
\SetAlgoLined
\textbf{Inputs:} $\bm{G},\{U\}, n_{goal}, r_s, \theta_s, l_s$  \\
\KwResult{Generate random goals within sensor range}
\While{$i \leq n_{goal}$}{ 
    $\rho^r$ = random\_point \\
    \If{\text{\textbf{not} is\_occupied}}{
        \For{j = 0, $n_u$}{
        
            \If{\text{in\_sensor\_view}}{
                \If{\text{\textbf{not} collision\_check}}{
                    $\bm{\rho}^r \gets [\bm{\rho}^r, \rho^r]$ \\
                    i++
                }
            }
        }
    }
}
\textbf{Output:} $\bm{\rho}_r$
\caption{Goal generation}\label{alg:goal_generation}
\end{algorithm}
where $n_u\in\mathbb{N}$ denotes the number of unknown voxels, and $\bm{\rho}^r \in \mathbb{R}^{n_{goal} \times 3}$ is the list of random goals. The result of this selection process is a list of candidate goals that the multi-goal RRT$^*$ can be tasked to plan potential paths to. In the evaluation of the algorithm we will set $r_s$ to \unit[6]{m}, a very conservative choice as high-end 3D lidar can have ranges of up to \unit[100]{m}, but it is generally good to under-value sensor ranges in mapping missions to guarantee detection hits. Moreover, $\theta_s$ and $l_s$ are set to \unit[32]{$^\circ$} and \unit[0.1]{m} approximating a Velodyne VLP16 Puck LITE. 
\subsubsection{Multi-goal RRT$^*$}
While there are numerous RRT implementations, ERRT uses a 3D RRT$^*$ structure. Random points $p^r$ are generated within $V_{map}$ and added as an end point $N_{end}$ to extend the closest graph vertex $N_{closest}$ in a graph network $\bm{N}$, if $p^r \in V_{free}$ and there is no collision/obstacles between the $p_r$ and $N_{closest}$. In RRT, this process continues until the goal $\rho$ is reached, generally as $\mid\mid \rho - N_{end} \mid\mid \leq d$, with $d$ denoting some specified distance. In RRT$^*$ the process is instead run for a pre-defined number of iterations, and the shortest generated graph that reached the goal (by the goal condition being satisfied), $N_{goal} \in \bm{N}$ is selected. The ERRT addition to this baseline framework of RRT algorithms, is the consideration of multiple goals, $\bm{\rho}^r$ provided by Algorithm 1. This is quite effective in the RRT$^*$ framework, as the graphs can be built as before, while only adding the evaluation of the goal conditions for each $\rho^r \in \bm{\rho}^r$, and after a specified number of iterations extracting the shortest graphs to each goal, which we can denote as $\bm{N}_{goal} \in \bm{N}$. Of course, for a limited number of iterations, there is no guarantee that a path can be found to each goal, but running the algorithm with a sufficiently high number of goals, the effect of discarding some goals is reduced. ERRT also adds logical operations to remove some unnecessary vertices in each selected graph, such as for each $N_{goal}$ connecting $\rho^r$ to the first collision-free vertex in $N_{goal}$ and discarding the rest, and removing redundant vertices in the graph. The resulting trajectories are interpolated to have a specified distance between points. The output of the multi-goal RRT$^*$ with some trajectory improvements is thus the shortest computed trajectory to each discovered goal, $\bm{x}_i$, with the non-discovered goals resulting in empty entries. 
As the 3D lidar field-of-view is not orientation-specific, the RRT$^*$ implementation considers the vehicle as a point-like object, and includes only position states. However, in the ERRT demonstration in Section IV, where an UAV model is used in \ref{sec:nmpc}, we consider the UAV as $x = [p,v,\phi,\theta]$ where $\phi$ and $\theta$ are the roll/pitch angles of the UAV, let us directly assume trajectory $\bm{x}$ consists of entries as $x = [p_x, p_y, p_z,0,0,0,0,0]$ e.g. only the position-states of the trajectory are set. 

\subsubsection{Trajectory Actuation via NMPC}\label{sec:nmpc}
Many exploration-planning frameworks are lacking a consideration of the predicted actuation a vehicle will use, while following a generated path. Minimizing actuation is a key component in limiting the energy utilization, while providing easier-to-follow (or more feasible) trajectories. Especially when the ERRT is considered for an UAV case a lower-actuation trajectory implies more stable flying behavior, which stresses localization, mapping, or detection software less as there are less rapid movements. Also, rovers or legged robots often traverse  slowly or need extra maneuvering to make tight turns that is not at all included in the path selection if the predicted model-based actuation along the path is not considered. In short: only evaluating information gain and the length of the trajectory (ex. \cite{bircher2016receding}) in trajectory selection misses key aspects that are more properly considered by also evaluating the predicted actuation along the trajectory. 

In general, there will be differences between the predicted actuation and the real live actuation of the vehicle, but for comparison any computation of information gain is similarly just a gauge for the approximate information gain as there is no way to know, for example, if more unknown voxels are right behind the frontier and would also be discovered. Similarly, solving a receding horizon NMPC problem where the reference set-points along the prediction horizon are the points in the generated trajectory, will result in the \textit{optimal} actuation based on the provided nonlinear dynamic model of the system with added constraints, and provides an approximation of the required actuation of the real vehicle (or rather the minimal required actuation). The flexibility in defining the NMPC cost function and constraints also allow platform-specific penalties when computing the actuation cost $J_a(\bm{u})$.

The NMPC approach used in ERRT follows previous works in the literature closely, while the same nonlinear UAV model is used\cite{small2019aerial, lindqvist2020nonlinear, sopasakis2020open}, with inputs as $u = [T, \theta_{ref}, \phi_{ref}]$ being references in thrust, roll, and pitch angles, but with a different application in this case. Thus, instead of solving the NMPC problem, with a very short sampling time to compute real-time control signals to a platform, we use a longer sampling time, combined with the interpolation length of trajectory $\bm{x}_i$, to match a desired predicted behavior and velocity of the vehicle. Let us denote $n_p\in\mathbb{N^+}$ as the prediction horizon, and $k+j|k$ the predicted time step $k+j$ produced at time step $k\in\mathbb{N^+}$. Similarly let $j\in\mathbb{N^+}$ also index the first $n_p$ entries in the path $\bm{x}$. As an example and based on the UAV case, the cost function, to make each predicted state reach the desired reference set as the entries in trajectory $\bm{x}$, and at each predicted time step, is:
\begin{multline}
\label{eq:costfunction}
J_{nmpc}(\bm{x}_{p,k}, \bm{u}_{k}, u_{k-1\mid k}) = \sum_{j=0}^{N-1} \big(  \smashunderbracket{\| x_j-x_{p,k+j{}\mid{}k}\|_{Q_x}^2}{\text{State penalty}}
\\
+   \smashunderbracket{\| u_{\mathrm{ref}}-u_{k+j{}\mid{}k}\|^2_{Q_u}}{\text{Input penalty}}
+  \smashunderbracket{\| u_{k+j{}\mid{}k}-u_{k+j-1{}\mid{}k} \|^2 _{Q_{\Delta u}}}{\text{Input change penalty}}\big) \\[1.2em]
\end{multline}
 where $Q_x, Q_t \in \mathbb{R}^{8\times8}, Q_u, Q_{\Delta u}\in 
\mathbb{R}^{3\times3}$ are positive definite weight matrices for the
states, inputs and input change respectively, $x_{p,k+j{}\mid{}k}$ are the predicted states and $u_{\mathrm{ref}}$ is the reference input, commonly set as a steady-state input of the platform. 
Let us also define input constraints as $u_{\min} \leq u_{k+j\mid k} \leq u_{\max}$. This leads to the following optimization problem:
\begin{subequations}\label{eq:nmpc}
\begin{align}
    \operatorname*{Minimize}_{
        \bm{u}_k, \bm{x}_{p,k}
    } \,
    & J_{nmpc}(\bm{x}_{p,k}, \bm{u}_{k}, u_{k-1\mid k})
    \\
    \text{subj. to:}\,& \notag
    x_{p,k+j+1\mid k} {=} \zeta(x_{p,k+j\mid k}, u_{k+j\mid k}), j{\in}\N_{[0, n_p-1]}, \notag
    \\
    &u_{\min} \leq u_{k+j\mid k} \leq u_{\max}, j\in\N_{[0, n_p-1]}, \notag
    \\
    &x_{p,k\mid k} {}={} \hat{x}_k, \notag
\end{align}
\end{subequations}
with $\zeta(x_{p,k+j\mid k}, u_{k+j\mid k})$ defining the discrete state model of the platform.
The optimization problem is solved, for each generated trajectory from the multi-goal RRT$^*$, to compute $\bm{u}_i$ with the Optimization Engine~\cite{sopasakis2020open}, following a \textit{single-shooting} approach. In the following evaluation we solve the problem with $n_p = 50$. Since the actuation is solved as a receding horizon problem, only $n_p$ predicted time steps can be considered and as such, we are solving for the actuation $\bm{u}$ only for the $n_p$ first entries in $\bm{x}$. In the case where the trajectory length is lower, the last entry is simply repeated. How far into the trajectory this limit of only considering $n_p$ entries is, depends on the interpolation of $\bm{x}_i$ and the sampling time $T_s$ of the NMPC. For the following evaluation case we use a sampling time of \unit[0.5]{s} and an interpolation length \unit[0.75]{m}, meaning that the desired platform's velocity to predict and optimize actuation for is \unit[1.5]{$\frac{m}{s}$}. 
%at a time, but the problem can be re-solved by simply initiating the NMPC with the last predicted state $x_{p,k+N{}\mid{}k}$ instead of the measured state $\hat{x}$ if the length of the trajectory exceeds $N$ to compute the actuation along the rest of the trajectory. 
We should also note that based on the optimal actuation vector $\bm{u}$ and the initial measured state vector $\hat{x}$, we can compute the full-state dynamic trajectory $\bm{x}_p$ up until the $N$:th entry, although despite the NMPC trajectory reference tracking, there is no guarantee that $\bm{x}_p$ is obstacle-free until the obstacle avoidance, based on the known map, is integrated into the NMPC framework. As such, in the following evaluation, we shall stick with a position-trajectory $\bm{x}$ to guarantee completely obstacle-free paths and only use the NMPC module solution to calculate the predicted actuation cost $J_a(\bm{u})$ along the trajectories, defined by the last two terms in \ref{eq:costfunction}, the input cost and the input rate cost. 

\subsubsection{Cost Calculation}
The final step of the ERRT algorithm is to evaluate each computed path in accordance with \eqref{eq:approx}. By denoting the number of entries in $\bm{x}_i$ as $n_i$, the distance cost of each trajectory is easily computed as the sum:
\begin{equation}\label{distcost}
   J_d(\bm{x}_i)_i = K_d \sum_{j=1}^{n_i} \mid \mid p_{i,j} - p_{i,j-1}\mid \mid.
\end{equation}
Based on the predicted actuation $\bm{u}_i$ along the trajectory, the actuation cost can be computed by feeding the actuation vector back into the relevant parts of the NMPC cost function. For the presented case it is
\begin{multline}
\label{eq:actuationcost}
J_{a}(\bm{u}_{i})_i = \sum_{j=1}^{n_p} \| u_{\mathrm{ref}}-u_{i,j}\|^2_{Q_u} 
+  \| u_{i,j}-u_{i,j-1} \|^2 _{Q_{\Delta u}}
\end{multline}
with $u_{\mathrm{ref}} = [9.81, 0, 0]$, the input that describes no movement for the UAV, with the thrust value of $9.81$ compensating for gravity. The total information gain $\nu$ is calculated by evaluating the information gain, at each point in the trajectory, as each unknown voxel within sensor view, and without obstructed line-of-sight and removing any duplicates (voxels seen at multiple points in the trajectory). For clarity the process is seen in Algorithm 2. 
\begin{algorithm}[htbp] 
\SetAlgoLined
\textbf{Inputs:} $\bm{G},\{U\},\bm{x}_i, n_{goal}, r_s, \theta_s, l_s$  \\
\KwResult{Information gain along the trajectory $\bm{x}_i$} 
    \For{j = 0, $n_i$}{ 
        \For{k = 0, $n_u$}{ 
            \If{\text{in\_sensor\_view}}{ 
                \If{\text{\textbf{not} collision\_check}}{
                    \text{seen\_unknown} $\gets$ [\text{seen\_unknown}, $U_k$]
                }
            }
        }
    }
$\nu$ = length(remove\_duplicates(\text{seen\_unknown))} \\
\textbf{Output:} $\nu$  
\caption{Information Gain}\label{alg:infogain}
\end{algorithm}

Once $\nu$ is computed, as the total number of unknowns that will be in sensor range by following $\bm{x}_i$, the exploration cost is then computed as: 
\begin{equation}
    J_e(\nu) = -K_\nu\nu
\end{equation}
with $K_\nu$ denoting a gain representing the relative emphasis on maximizing the information gain. With $J_a(\bm{u_i})_i, J_e(\nu_i)_i, J_d(\bm{x}_i)_i$ calculated for $i = 0,\ldots n_{goal}$ \eqref{eq:approx} can be evaluated and the $\bm{x}_i$ related to the minimum-cost solution and denoted by $\bm{x}_\mathrm{min}$ is selected as the final result of the algorithm.

\subsubsection{ERRT input structure}
The overall ERRT framework has been implemented in ROS~\cite{quigley2009ros}, with a custom message including all input parameters to the algorithm, which are summarized in Table \ref{table:inputs}.  \\

\begin{table}[ht]
\begin{tabularx}{\columnwidth} { 
  | >{\raggedright\arraybackslash}X 
  | >{\centering\arraybackslash}X 
  | >{\raggedleft\arraybackslash}X | }
 \hline
 Occupied voxels & $\{O\}$  \\
 \hline
 Unknown voxels & $\{U\}$  \\
\hline
 Vehicle state  & $\hat{x}$  \\
\hline
 Number of goals  & $n_{goal}$  \\
\hline
 Sensor model parameters & $r_s, \theta_s, l_s$  \\
\hline
 Cost parameters & $K_d, K_\nu, Q_u, Q_{\Delta u}$  \\
\hline
RRT$^*$ iterations & $iter$  \\
\hline
Grid resolution & $g_{res}$  \\
\hline
\end{tabularx}
\caption{ERRT Initialization Parameters.}\label{table:inputs}
\end{table}
At every call to the algorithm the grid map \textbf{G} is initialized based on point clouds of occupied and unknown voxels, and the user is free to change resolution, size of the map, or any other parameter at every call to the algorithm to meet the desired mission specifications. After calculation is complete the node publishes the computed trajectory. The very general input model was one of the motivations for developing ERRT, allowing for example high-resolution exploration of the local area, and low-resolution computation of what area to explore next, by the same planner with different inputs. The process described in Section \ref{sec:methodology} can be summarized in Algorithm 3. 

\begin{algorithm}[ht] 
\SetAlgoLined
\textbf{Inputs:} $\bm{G},\{U\}, n_{goal}, r_s, \theta_s, l_s,\hat{x}$  \\
\KwResult{Minimum-cost trajectory $\bm{x}_{min}$} 
$\bm{\rho}^r$ = \text{generate\_goals}$(\bm{G},\{U\}, n_{goal}, r_s, \theta_s, l_s)$ \\
$\bm{N}_{goal}$ = \text{multigoal\_rrt}$(\bm{\rho}^r, \bm{G})$ \\
$\bm{x}_i$ = \text{trajectory\_improvement}$(\bm{\rho}^r,\bm{N}_{goal} ,\bm{G})$ \\
$\bm{u}_i$ = \text{NMPC\_module}$(\bm{x}_i, \hat{x})$ \\
$(J_a, J_d, J_e)_i$ = \text{cost\_calc}$(\{U\},\bm{G}, \bm{x}_i, \bm{u}_i, K_d, K_\nu, Q_u, Q_{\Delta u})$ \\
$\bm{x}_{min}$ = \text{path\_selection}$((J_a, J_d, J_e)_i, \bm{x}_i)$ \\
\textbf{Output:} $\bm{x}_{min}$  
\caption{The ERRT algorithm.}\label{alg:errt}
\end{algorithm}
%%%%%%%%%%%%%%%%%%%%%%%%%%%%%%%%%%%%%%%%%%%%%%
%%%%%%%%%%%%%%%%%%%%%%%%%%%%%%%%%%%%%%%%%%%%%%
\section{Results} \label{sec:results}
The proposed method is evaluated in a simulated environment separated from other mapping or frontier-generating software, that considers a nonlinear dynamic UAV model with added small magnitude localization noise, and a sensor model that mimics the model described in \ref{sec:goals} with a \unit[1]{m} longer sensor range for a voxel to be considered discovered than what is considered in the ERRT. The map is initialized with the full space as undiscovered, except a small area around the starting location, and as the center of a voxel comes into sensor range, it is set either as free space or as occupied. As the planner is ROS-integrated, the exploration process can be easily visualized in Rviz\cite{kam2015rviz}. It should be noted that for the sake of visualization, the occupied voxels are always depicted. A full-state reference tracking controller, tuned to approximately match the desired predicted behavior as stated in \ref{sec:nmpc}, follows the generated path, until the goal is reached and the path is recalculated. 

\begin{figure}[ht!]
    \centering
\includegraphics[width=\columnwidth]{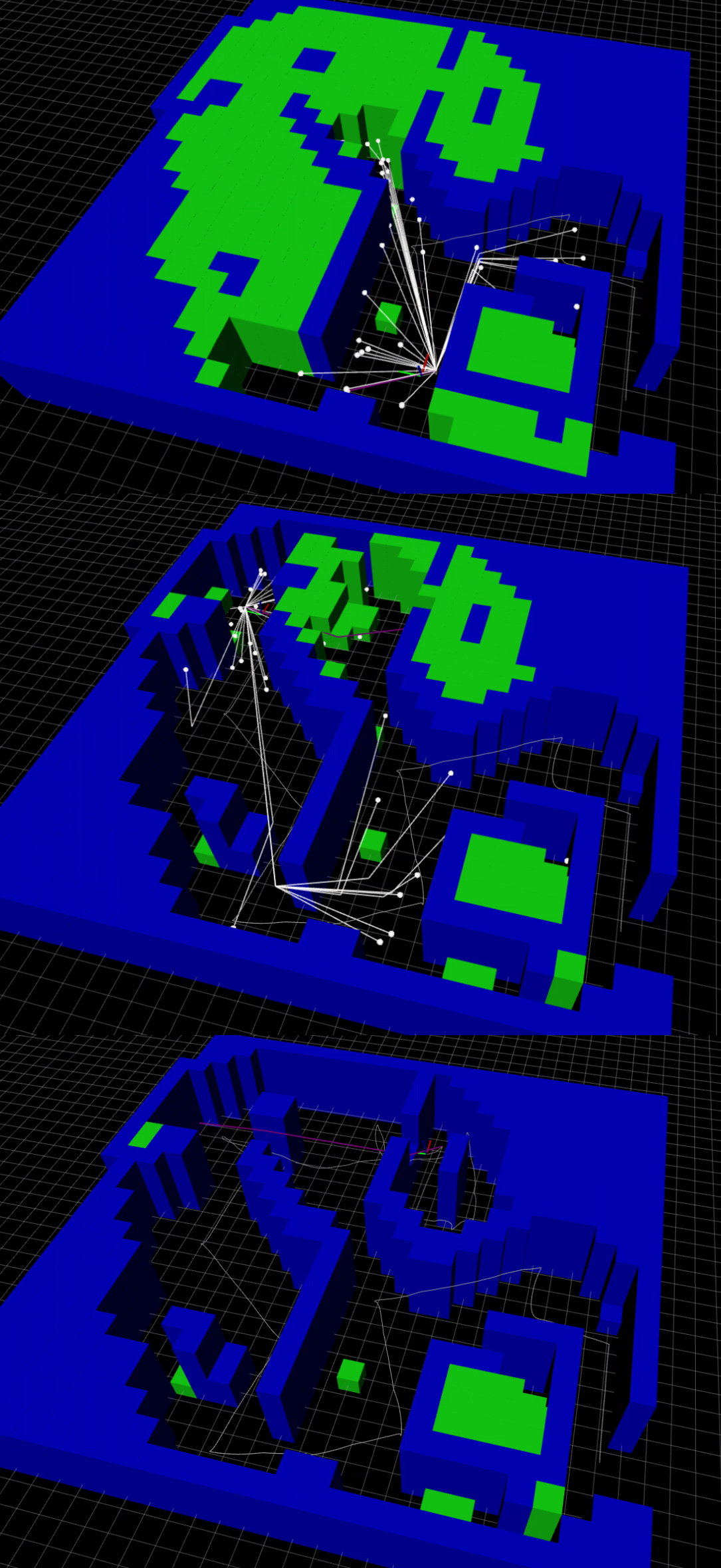}
    \caption{Selected point during exploration. Unknown areas (green voxels), all generated paths (white), selected (lowest cost) path (purple), total exploration path (grey). a) greedily maximizing information gain, b) selecting to ignore some unexplored voxels to go to the more information-dense area, c) cleaning up after exploring high-information open-areas.}
    \label{fig:exploration}
\end{figure}

\begin{figure}[ht]
    \centering
\includegraphics[width=0.9\columnwidth]{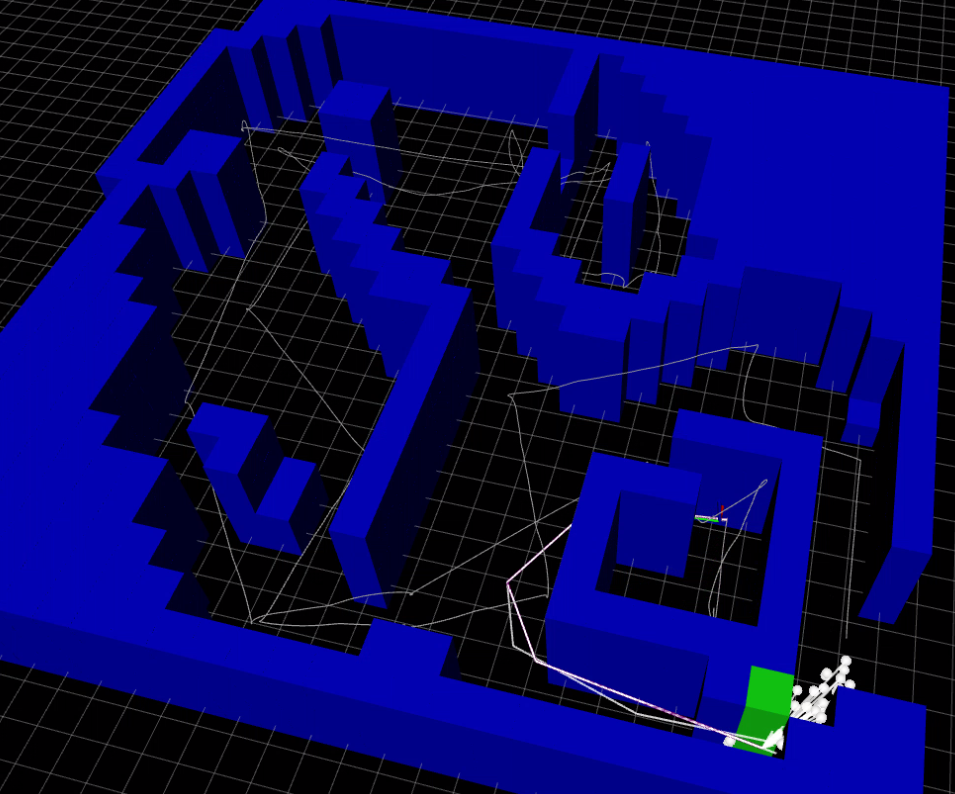}
    \caption{Exploring the last unknown area.}
    \label{fig:greedy_exp_6}
\end{figure}

The considered evaluation environment has been generated to mimic a subterranean cave area with multiple rooms, a larger void area, and several small hard-to-reach nooks and passages, that encompasses a total of \unit[27x27x4]{m$^3$}, with a grid size of \unit[1]{m}. The ERRT algorithm was executed for a total of twenty times in the simulated area, ten times with a greedy tuning prioritizing information gain over minimizing distance and actuation and a more conservative tuning with higher costs on distance and actuation as compared to information-gain. Figures \ref{fig:exploration} and \ref{fig:greedy_exp_6} present time sampled images through the evolution of the mission from one of the greedy runs, displaying all the generated paths at critical times during the simulation. We evaluate the simulations in terms of the total distance travelled, and the time required to complete the task, considering both at 90\% of voxels discovered as a gauge for the effective volumetric gain, and at full completion. 
\begin{figure}[ht]
    \centering
\includegraphics[width=\columnwidth]{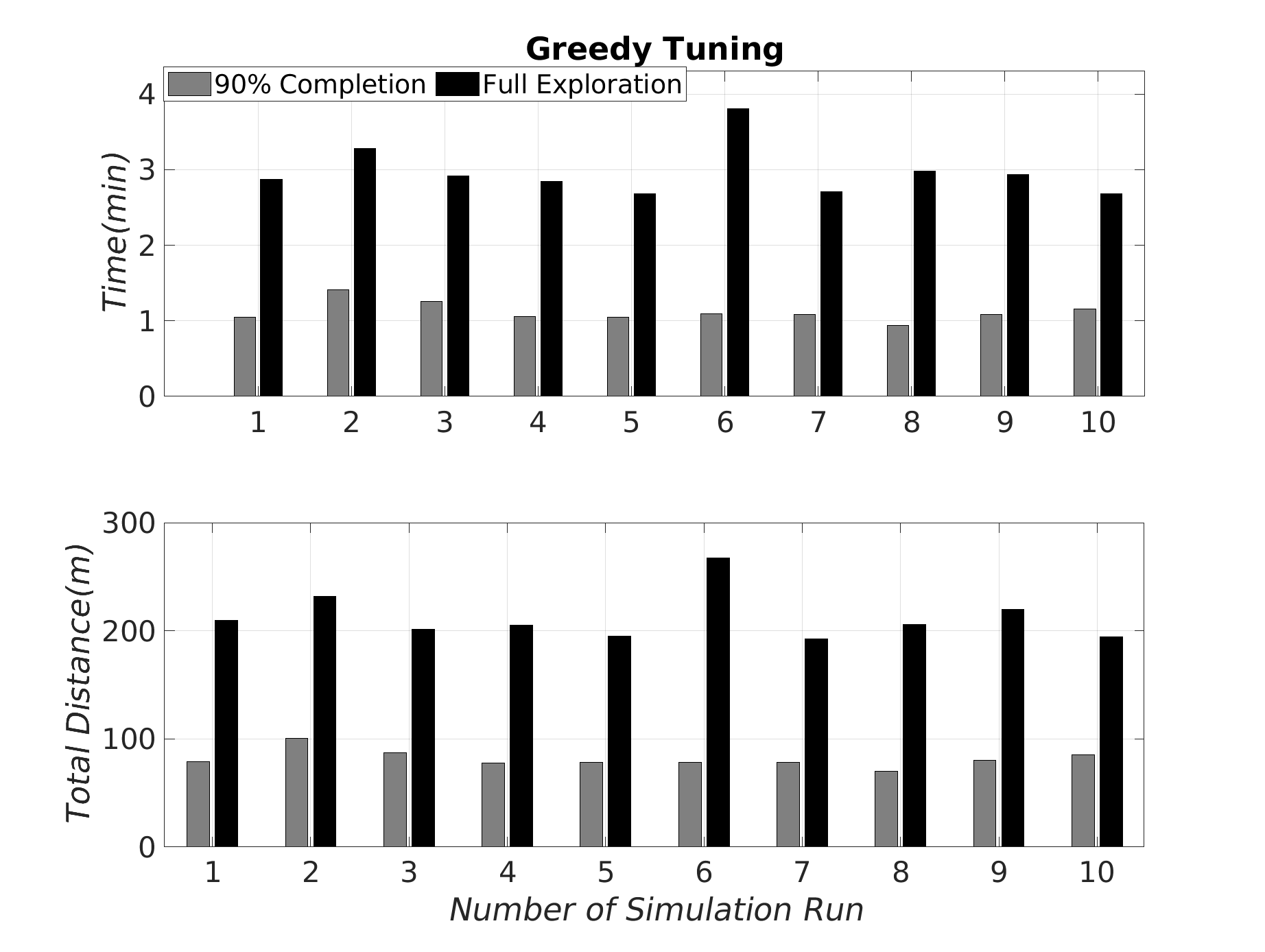}
    \caption{Exploration time and total distance travelled for the greedy tuning of Exploration-RRT.}
    \label{fig:greedy}
\end{figure}

\begin{figure}[ht]
    \centering
\includegraphics[width=\columnwidth]{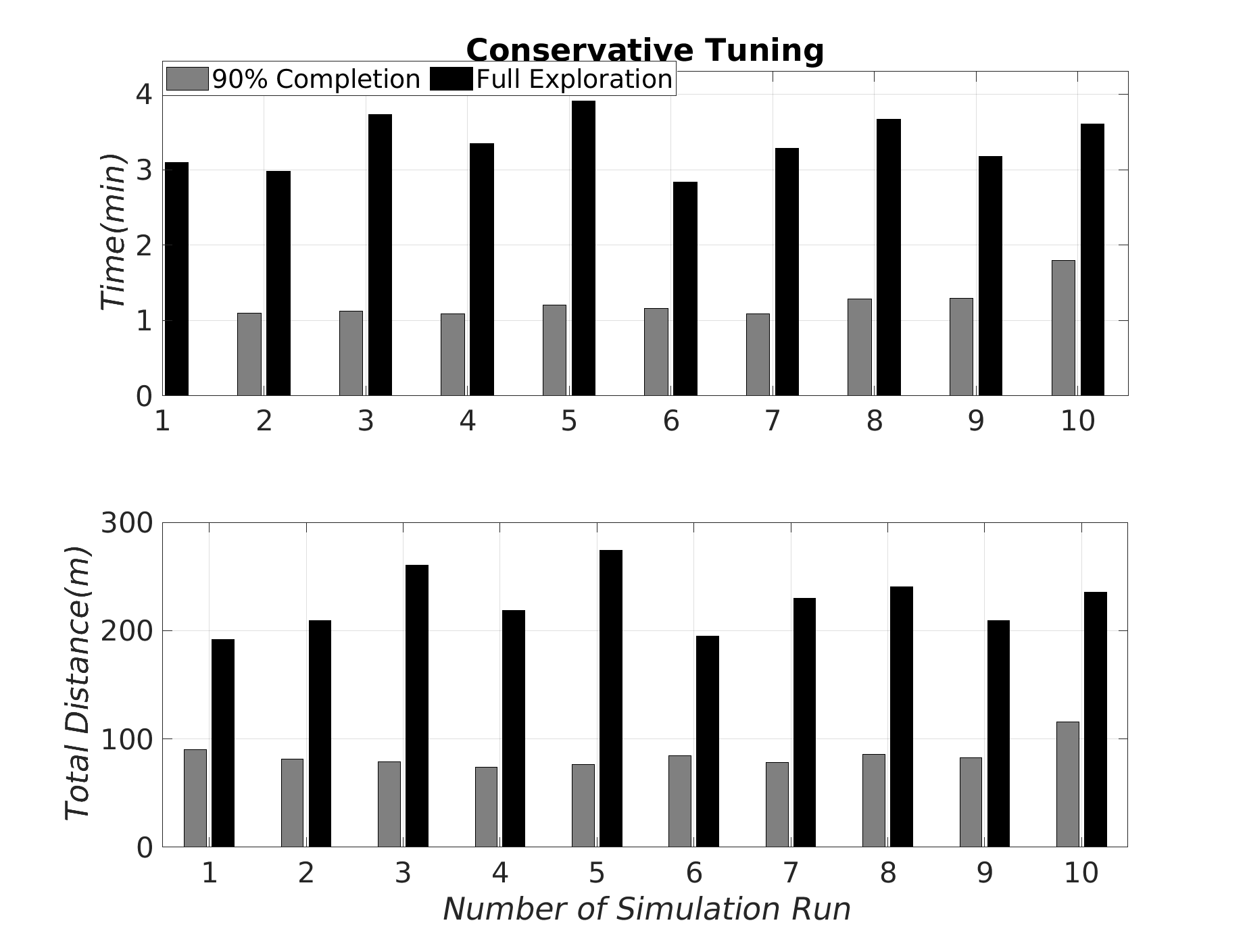}
    \caption{Exploration time and total distance travelled for the conservative tuning of Exploration-RRT.}
    \label{fig:convervative}
\end{figure}

\begin{figure}[ht]
    \centering
\includegraphics[width=0.8\columnwidth]{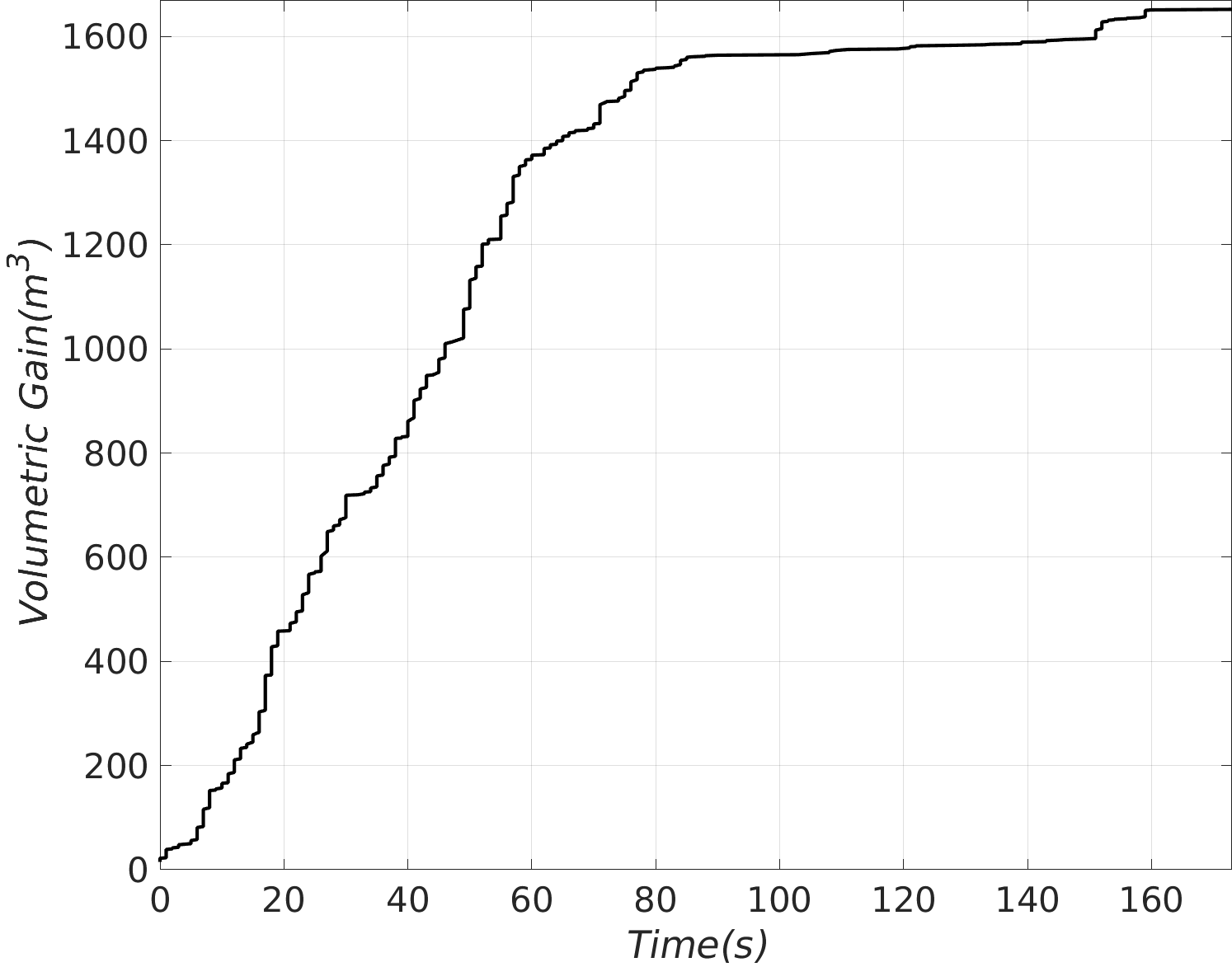}
    \caption{Volumetric Gain from one of the greedy runs.}
    \label{fig:infogain}
\end{figure}

The results for the greedy tuning can be found in Figure \ref{fig:greedy} and for the conservative in Figure \ref{fig:convervative}. As it can be observed, a majority of the exploration time is devoted to cleaning up voxels that were ignored along the way due to maximizing the information gain, and exploring the final hard to reach areas to achieve complete coverage of the area. In Figure \ref{fig:infogain} the volumetric gain (or information gain) from one of the runs is depicted, which also represents the behavior of maximizing the total explored volume as quickly as possible. Both configurations show very consistent results over multiple executions, despite the very different behavior for each tuning, with the greedy tuning over-all performing better. This is naturally the case in a completely unknown area, as maximizing the information gain is of paramount importance and the exploration missions are often time-constrained by the platform that further promotes the greedy behavior. The average translation speed of the platform was \unit[1.2]{m/s}, controlled by the interpolation of the trajectories, and the tuning of the reference tracking controller, while mimicking the slower pace of a fully autonomous UAV in a subterranean field application. A video compilation of different exploration executions in the evaluation environment can be found here: \url{https://drive.google.com/file/d/1v3vg3Z9iB2DR-Oec3MxWUg_39d3lIj1F/view?usp=sharing}. 

The ERRT algorithm was running 1500 $\text{RRT}^*$ iterations and the $n_{goal}$ was set to 40, which resulted in solver times of \unit[0.3-0.9]{s}, averaging at \unit[0.6]{s}, of which around \unit[0.1]{s} was consumed by the NMPC module (around \unit[2.5]{ms} per goal). Although the framework is set up as to optimize over the full trajectory, it needs to be speed up to enable an online implementation where trajectories are recalculated at each execution step, mainly related to the multi-goal RRT$^*$. 

%\begin{figure}[!htb]
%    \centering
%\includegraphics[width=\columnwidth]{figures/environment.%png}
%    \caption{A generated subterranean-like environment to test the algorithm with multiple rooms, a larger void, and small nooks.}
%    \label{fig:env}
%\end{figure}

\section{Future Work} \label{sec:futurework}
From the algorithmic perspective, there are two main directions of future work: speeding up the core multi-goal RRT$^*$ framework which would expand the planner's application areas, and adding integrated collision avoidance that considers the size-radius of the platform both in terms of obstacle enlargement in the main RRT$^*$ planner and in the NMPC optimization problem to be able to guarantee the actuator-based paths $\bm{x}_p$ are completely obstacle free. 
Another interesting direction is adding negative costs related to moving through areas with well-defined features, which are commonly tracked by SLAM software~\cite{shan2020lio}, as doing so could improve localization accuracy. This would allow the planner to also consider the \textit{localization cost}, and can be implemented into the framework in a very straight-forward way assuming that the locations of such features is provided by the SLAM software. 
From the implementation perspective we aim to release the framework as an open-source ROS package, and evaluate it with mapping, frontier-generation, and reactive obstacle avoidance software in the loop. 

%%%%%%%%%%%%%%%%%%%%%%%%%%%%%%%%%%%%%%%%%%%%%%%%%%%%%%%%%%%%
\section{Conclusions} \label{sec:conclusion}
This paper has presented a novel algorithm for combined exploration and path planning behavior towards the goal of considering the path planning and information-gain (exploration) as a coupled problem. In the selected initial evaluation environment, the algorithm quickly and efficiently explored all unknown areas of the map, showing a consistent behavior over multiple exploration runs and without failure to complete the task of complete coverage.
%%%%%%%%%%%%%%%%%%%%%%%%%%%%%%%%%%%%%%%%%%%%%%%%%%%%%%%%%%%%

\bibliography{mybib}
\end{document}